\documentclass[sigconf]{acmart}

\AtBeginDocument{%
  \providecommand\BibTeX{{%
    \normalfont B\kern-0.5em{\scshape i\kern-0.25em b}\kern-0.8em\TeX}}}

\setcopyright{acmcopyright}
\copyrightyear{2021}
\acmYear{2021}
\acmDOI{}

\usepackage{multirow}
\usepackage{CJK}
\usepackage[T1]{fontenc}
\usepackage{float}

\begin{document}

\title{CODER: Knowledge infused cross-lingual medical term embedding for term normalization}

\author{Zheng Yuan, Zhengyun Zhao}
\email{{yuanz17,zhao-zy15}@mails.tsinghua.edu.cn}
\affiliation{%
  \institution{Tsinghua University}
  \streetaddress{}
  \city{Beijing}
  \country{China}
}

\author{Haixia Sun, Jiao Li}
\email{{sun.haixia,li.jiao}@imicams.ac.cn}
\affiliation{%
  \institution{Chinese Academy of Medical Sciences / Peking Union Medical College}
  \streetaddress{}
  \city{Beijing}
  \country{China}}

\author{Fei Wang}
\email{few2001@med.cornell.edu}
\affiliation{%
 \institution{Cornell University}
 \streetaddress{}
 \city{New York}
 \state{NY}
 \country{USA}}

\author{Sheng Yu}
\authornote{Corresponding author.}
\email{syu@tsinghua.edu.cn}
\affiliation{%
 \institution{Tsinghua University}
 \streetaddress{}
 \city{Beijing}
 \state{}
 \country{China}}

\renewcommand{\shortauthors}{Yuan and Zhao, et al.}

\begin{abstract}
This paper proposes CODER: \textbf{co}ntrastive learning on knowle\textbf{d}ge graphs for cross-lingual m\textbf{e}dical term \textbf{r}eprensentation.
CODER is designed for medical term normalization by providing close vector representations for different terms that represent the same or similar medical concepts with cross-lingual support.
We train CODER via contrastive learning on a medical knowledge graph (KG) named the Unified Medical Language System \cite{Bodenreider2004}, where similarities are calculated utilizing both terms and relation triplets from KG.
Training with relations injects medical knowledge into embeddings and aims to provide potentially better machine learning features.
We evaluate CODER in zero-shot term normalization, semantic similarity, and relation classification benchmarks, which show that CODER outperforms various state-of-the-art biomedical word embeddings, concept embeddings, and contextual embeddings.
Our codes and models are available at \url{https://github.com/GanjinZero/CODER}.
\end{abstract}

\ccsdesc[500]{Computing methodologies~Knowledge representation and reasoning}


\keywords{medical term normalization, cross-lingual, medical term representation, knowledge graph embedding, contrastive learning}


\maketitle

\raggedbottom
\showboxdepth=5
\showboxbreadth=5

\section{Introduction}

Electronic medical records (EMRs) contain plentiful medical information that can assist various medical research, such as clinical decision support \cite{Miotto2016}, automated phenotyping \cite{yu2018enabling}, and medical knowledge mining \cite{Lin2020}.
However, analyzing the free text in EMRs remains challenging because the same medical concept can appear in various nonstandard names, abbreviations, and misspellings, which all need to be normalized to their corresponding standard terms or their concept IDs of an existing terminology system, such as the Unified Medical Language System (UMLS) \cite{Bodenreider2004} or SNOMED Clinical Terms (SNOMED-CT) \cite{donnelly2006snomed}.
Normalization is also important for structured EMR data.
Different medical institutes use different coding systems for laboratory examinations, procedures, and medications.
Thus, when conducting multi-institutional studies, normalization can facilitate mapping the database fields to the same standard for joint analysis.

Medical term normalization can be regarded as either classification or ranking problem.
Neural-network based techniques encode terms into hidden representations by recurrent neural networks or pre-trained language models (PLM).
A softmax layer on the hidden representations \cite{Limsopatham2016,Tutubalina2018,Niu2019,Miftakhutdinov} is used for classification.
Classification methods achieve state-of-the-art scores in benchmark tests but are limited to a small set of target concepts.
In practice, terminology systems in use, such as the UMLS, have millions of concepts.
Normalizing medical terms to these systems remains challenging for the classification approach due to the enormous number of classes and the inflexibility to add new classes.
Another approach is to rank candidate concepts by similarity to the term \cite{leaman2013dnorm,li2017cnn,sung-etal-2020-biomedical}.
It differs from the previous approach in that it is trained as a binary classifier, where terms and their corresponding concepts’ names (target terms) form positive samples, and terms and non-corresponding concepts’ names form negative samples.
The scalar output from the classifier is then used as a measure of similarity to rank the candidate concepts for normalization.
Similar to the classification approach, these models are trained on labeled datasets with limited target concepts.

To achieve term normalization on very large scale and flexible concept sets, we propose CODER based on PLM with KG contrastive learning.
CODER generates embeddings for terms and ranks the candidate concepts by the cosine similarity between the embeddings of the candidate target terms and that of the input term, avoiding training a classification model.
Existing medical embeddings can be classified into word embeddings \cite{moen2013distributional,Chiu2016,info:doi/10.2196/14499}, concept embeddings \cite{DeVine2014,Choi20161,Choi20162,cai2018medical,beam2018clinical,zhang2020learning}, and contextual embeddings \cite{huang2019clinicalbert,alsentzer2019publicly,beltagy-etal-2019-scibert,lee2020biobert,gu2020domain,peng2019transfer,jin2019probing,zhang2020learning}.
Word and concept embeddings have good base performances on evaluating similarity, but face the out-of-vocabulary (OOV) problem and cannot handle misspellings that are commonly present in clinical text.
PLM-based contextual embeddings can relieve the OOV problems by using sub-word tokenization, but they have weaker performance than word and concept embeddings on evaluating similarity if not fine-tuned \cite{reimers-gurevych-2019-sentence}.

Contrastive learning is a popular technique in computer vision (CV) \cite{moco,simclr} as a solution to learning representations with a very large number of classes.
In each training step, instead of learning to map samples to their labels, contrastive learning aims to learn sample representations to distinguish samples of the same label (positive pairs) from those of different labels (negative pairs).
For term normalization, terms of the same and different concepts naturally serve as positive and negative pairs in contrastive learning, which has been explored by SapBERT \cite{sapbert} using the UMLS terminology.
However, as a KG, the UMLS contains more knowledge than just terms, and knowledge can serve as useful information for normalization.
For example, when normalizing ``\textit{poisoned by eating pufferfish}'' we may want ``\textit{food poisoning}'' to rank among the top because pufferfish is a kind of food.
For another example, it would be more meaningful to have the embedding of ``\textit{rheumatoid arthritis}'' closer to ``\textit{osteoarthritis}'' than ``\textit{rheumatoid pleuritis}'' because both ``\textit{rheumatoid arthritis}'' and ``\textit{osteoarthritis}'' are subtypes of arthritis, which may affect the same body parts and share treatments.
The reasoning in these examples illustrates that relational knowledge is essential to achieve meaningful medical embeddings.
Therefore, we propose dual contrastive learning on both terms and relation triplets of KGs.
The term-relation-term similarity are measured between a term-relation $(h,r)$ and a term $(t)$.
Positive and negative term-relation-term pairs are determined by KGs.
Online hard pairs mining is applied to find informative pairs which are often used in person re-identification (ReID) \cite{reid}.
ReID and medical term normalization are similar tasks since they both have many categories and sparse samples for each category.

To support cross-lingual term normalization, we use mBERT \cite{devlin2018} which encodes texts in different languages to one unified space. 
Previous methods are usually translation-based and rely on biomedical parallel corpora \cite{Afzal2016,roller2018cross,perez2020cross}.
The UMLS naturally contains massive cross-lingual medical synonyms which have not been fully utilized.

The contributions of this paper include:

\begin{itemize}
    \item We propose a KG contrastive learning model for term normalization that uses both synonyms and  relations from the UMLS to direct the generation of medical term embeddings.
    The learning strategy can also be applied to other KGs.
    \item We perform evaluations on CODER against notable existing medical embeddings.
    CODER achieves state-of-the-art results in zero-shot term normalization, medical concept similarity measure, and concept relation classification tasks.
    CODER embeddings can be used for embedding-based term normalization directly or features for machine learning, and CODER can be fine-tuned like other PLMs.
    \item CODER is the first cross-lingual medical term representation with support for English, Czech, French, German, Italian, Japanese, Portuguese, Russian, Spanish, Dutch, and Chinese.
\end{itemize}




\section {Related work}
In this section, we briefly introduce the medical term normalization task, existing medical embedding, and contrastive learning.

\subsection{Medical term normalization}
Classification methods generate hidden representations of terms and use softmax layer to classify terms into concepts.
Terms are encoded by CNN, RNN, or PLM \cite{Limsopatham2016,Tutubalina2018,Miftakhutdinov}.
The attention mechanism is introduced to capture important words or characters for classification \cite{Tutubalina2018,Niu2019}.
These classification methods only consider target terms as a category and do not utilize semantic information from target terms.

Ranking methods rank the similarity between input term and candidate target terms by training them as positive and negative pairs.
DNorm \cite{leaman2013dnorm} learns to rank target terms via calculating similarities between TF-IDF vectors.
\citet{li2017cnn} encodes input and target terms by CNN and uses a pairwise approach for ranking.
NSEEN \cite{fakhraei2019nseen} trains a siamese LSTM network and uses hard negative samplings for finding informative pairs.
BNE \cite{phan2019robust} encodes terms, concepts, and contexts separately, and trains the model based on term-term, term-concept, and term-context similarities.
\citet{pattisapu2020medical} calculate graph embeddings for the target terms and rank them by the cosine distance.
BIOSYN \cite{sung-etal-2020-biomedical} uses both TF-IDF and BioBERT \cite{lee2020biobert} to represent terms. Synonym marginalization is used for maximizing similarities between synonyms. 

Non-English medical term normalization has also attracted the interest of researchers.
\citet{Niu2019} constructed the Chinese Medical Concepts Normalization dataset.
Translation-based methods have been used for French, Spanish, Dutch, and German \cite{Afzal2016,roller2018cross,perez2020cross}.

\subsection{Medical embeddings}

Word embeddings are typically trained by the word2vec model \cite{Mikolov2013} using medical records or papers \cite{moen2013distributional,Chiu2016,info:doi/10.2196/14499}.

Concept embeddings map medical concepts to embedding vectors and cannot perform term normalization intrinsically. 
Sequences of medical concepts identified from EMRs or biomedical papers are considered as sentences to train embeddings with word2vec \cite{DeVine2014,Choi20161,Choi20162,cai2018medical}.
Cui2vec \cite{beam2018clinical} factorizes the CUI-CUI pointwise mutual information matrix to obtain embeddings.
\citet{zhang2020learning} proposed Conceptual-Contextual embedding trained on both biomedical corpora and UMLS relations.


Contextual embeddings can generate different embeddings according to the context.
The popularity of contextual embeddings stems from large-scale PLM like BERT \cite{devlin2018} and ELMo \cite{peters2018deep}.
Several medical contextual embeddings have been trained on different medical corpora under the architecture of BERT \cite{huang2019clinicalbert,alsentzer2019publicly,beltagy-etal-2019-scibert,lee2020biobert,gu2020domain,peng2019transfer}.
\citet{jin2019probing} proposed BioELMo by training on PubMed.
SapBERT \cite{sapbert} conducts self-alignment pre-training on UMLS synonyms for term embeddings.
Contextual embeddings have great potential for various kinds of downstream NLP tasks but typically require supervised fine-tuning.
We found most existing contextual embeddings perform worse on zero-shot term normalization than word embeddings since the ${\rm [CLS]}$ token of masked language models may not reflect semantic similarity directly \cite{li-etal-2020-sentence}.

\subsection{Contrastive learning}
Contrastive learning has been successful in learning representations from the comparison between positive and negative pairs, especially in the field of CV \cite{contrastive-summary}.
SimCLR \cite{simclr}, MoCo \cite{moco} and SwAV \cite{swav} have achieved comparable results with supervised methods in the ImageNet dataset.
Data augmentation is required for self-supervised contrastive learning to generate positive pairs. 
Crop, Gaussian blur, and color distortion \cite{simclr} can be used in the CV domain.
For NLP, DeCLUTR \cite{declutr} collects positive sentence pairs from the same document. Back-translation is applied in CERT \cite{fang2020cert} to create similar sentence pairs.
SapBERT \cite{sapbert} utilizes synonyms from the UMLS as positive pairs.
Our model fully utilizes the UMLS as a KG and trains on both synonyms and relation triplets.
Loss functions are another essential components for contrastive learning to maximize similarities between positive pairs and minimize similarities between negative pairs.
Triplet loss \cite{triplet-loss} minimizes the relative distance between one positive and one negative pair.
InfoNCE \cite{oord2018representation} considers all other samples in a mini-batch as negative pairs.

\section{Method}

\begin{figure*}[h]
\centering
\includegraphics[scale=0.45]{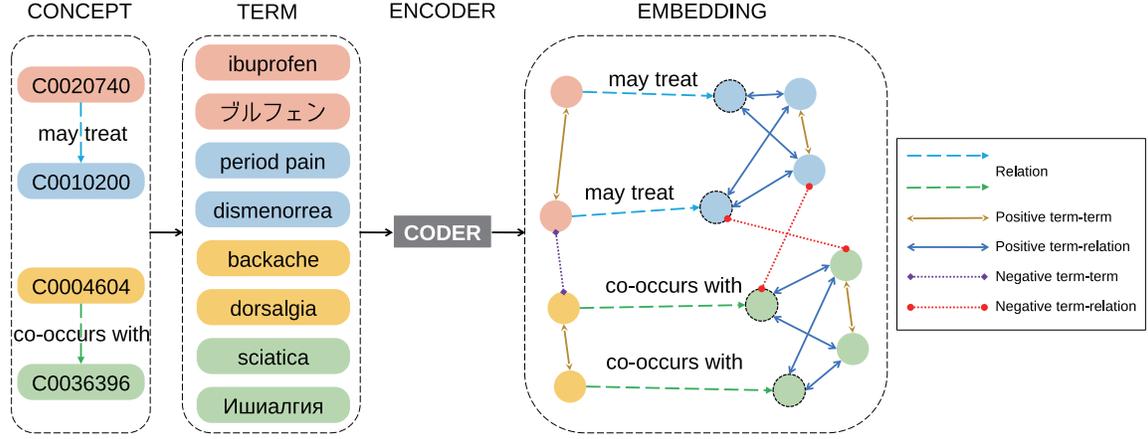}
\caption{\label{workflow} The overview of CODER. CODER encodes terms potentially in different languages into the embedding space. Term-term similarities and term-relation-term similarities are calculated to train CODER.}
\Description{There are four medical concepts C0020740, C0010200, C0004604 and C0036396 with two relation. Terms are in different languages. CODER encodes terms into embeddings. Term-term similarities and term-relation-term similarities are calculated to train CODER.}
\end{figure*}

In this section, we introduce the model architecture of CODER.
We first define the embedding-based term normalization task and how we embed terms.
Then we introduce the KG contrastive learning for term normalization.


\subsection{Embedding-based Term Normalization}
Let $\mathcal{D}=\{n_i\}_{i=1}^{\|\mathcal{D}\|}$ be a concept dictionary where $n_i$ is a concept.
Each concept $n_i$ has several terms $\{s_i^j\}_{j=1}^{c_i}$ as synonyms.
We define the embedding assigned to term $s_i^j$ as $\mathbf{e}_i^j \in \mathbb{R}^l$.
For an input string $s$ with embedding $\mathbf{e}$, the goal is to predict its concept $n$:
\[
    \hat{n} = n_{\mathop{\arg\max}_i(\cos(\mathbf{e}_i^j,\mathbf{e}))},
\]
where $\cos$ is the cosine similarity measure in Euclid space.

\subsection{Term representation}\label{rep}
CODER takes an input term $s$ and outputs an embedding vector $\mathbf{e}$.
The input term $s$ is tokenized to sub-words ${\rm [CLS]}, s_0, s_1, ..., s_k, {\rm [SEP]}$ with two special tokens.
A pre-trained language model $\rm PLM$ encodes $s$ to a series of hidden states $\mathbf{h}_i$: 
\[
    \mathbf{h}_{\rm [CLS]}, \mathbf{h}_0, ..., \mathbf{h}_k, \mathbf{h}_{\rm [SEP]} = {\rm PLM}({\rm [CLS]}, s_0, s_1, ..., s_k, {\rm [SEP]}).
\]
We initialize the $\rm PLM$ by PubMedBERT \cite{gu2020domain} or mBERT \cite{devlin2018}.
The embedding of $s$ is taken from the representation of the {\rm [CLS]} token:
\[
    \mathbf{e} = \mathbf{h}_{\rm [CLS]}.
\]
We can also embed the term by average pooling:
\[
    \mathbf{e}_{avg}={\rm AvgPool}(\mathbf{h}_{\rm [CLS]}, \mathbf{h}_0, ..., \mathbf{h}_k, \mathbf{h}_{\rm [SEP]}).
\]

\subsection{The contrastive learning framework for KG with synonyms} \label{method}

CODER learns term representations by maximizing similarities between positive term-term pairs and term-relation-term pairs from a KG.
The KG records relation triplets $\{(h,r,t)\}$, where the head concept $h$ and the tail concept $t$ are from concept dictionary $\mathcal{D}$, and $r\in\mathcal{R}$ is the relation between them.
At each training step, we sample a batch of relation triplets $\{(h_i,r_i,t_i)\}_{i=1}^k$ from the KG.
For convenience of notation, denote $n_i=h_i, n_{k+i}=t_i, \forall 1 \leq i \leq k$.
We sample term $s_i$ for each $n_i$ from $\{s_i^j\}_{j=1}^{c_i}$ in $\mathcal{D}$.
CODER embeds terms $\{s_i\}_{i=1}^{2k}$ into $\{\mathbf{e}_{i}\}_{i=1}^{2k}$.

We define the term-term pair label $\tau_{ij}$ in a mini-batch based on the concepts:
\[
    \tau_{ij} = \left\{
\begin{aligned}
1&, & n_i=n_j, \\
0&, & n_i\neq n_j.
\end{aligned}
\right.
\]
The similarity between two terms in a mini-batch is defined as:
\[
    S_{ij} = \cos(\mathbf{e}_i, \mathbf{e}_j).
\]

A series of pair-based loss functions, such as the InfoNCE \cite{oord2018representation} have been proposed for learning representations in contrastive learning.
We adopt the Multi-Similarity loss (MS-loss) \cite{wang2019multi} to learn embeddings, which incorporates hard pair mining in its formulation to accelerate the convergence.
Other pair-based losses can also be used as the loss function, and we leave the comparative study of loss functions to future work.
The MS-loss mines hard pairs according to positive relative similarity.
For an anchor $i$, it defines:
\[
    \mathscr{N}_i\coloneqq\{j\vert\tau_{ij}=0,S_{ij}>\min_{\tau_{ik}=1}S_{ik}-\epsilon\},
\]
\[
    \mathscr{P}_i\coloneqq\{j\vert\tau_{ij}=1,S_{ij}<\max_{\tau_{ik}=0}S_{ik}+\epsilon\},
\]
where $\epsilon$ is a margin.
The MS-loss of a mini-batch is:
\[
    \mathcal{L}_{MS} = \frac{1}{2k}\sum_{i=1}^{2k}(\frac{\log(1+\sum_{j\in\mathscr{P}_i}\exp(-\alpha(S_{ij}-\lambda)))}{\alpha}
\]
\[
    +\frac{\log(1+\sum_{j\in\mathscr{N}_i}\exp(\beta(S_{ij}-\lambda)))}{\beta}),
\]
where $\alpha, \beta, \lambda$ are hyper-parameters.
$\mathscr{P}_i$ will vanish if $\tau_{ij}=0, \forall j\neq i$, and $\mathscr{N}_i$ will vanish if $\tau_{ij}=1, \forall j\neq i$.


Toward injecting relational knowledge from KG into term representations, CODER also learns KG embedding inspired by semantic matching methods like DistMult \cite{distmult} and ANALOGY \cite{analogy}, which aim to approximate $M_r^\top h \approx t$.
The motivation is to learn a better similarity function via relations.
Consider two triplets $(h_0,r,t_0)$ and $(h_1,r,t_1)$ with the same relation type, $h_0$ and $h_1$ being semantically similar (e.g. similar diseases) may suggest that $t_0$ and $t_1$ are also semantically similar (e.g. similar drugs).
$M_r^\top h \approx t$ will guarantee the relation-based similarity inference in the above example.
$(h_i,r_i)$ can be viewed as a relation-based data augmentation of $t_i$.
We can define the term-relation-term similarity between relation $(s_i, r_i), \forall 1 \leq i \leq k$ and term $s_j, \forall k + 1 \leq j \leq 2k$ in a mini-batch by:
\[
    S^{rel}_{ij} = \cos(\mathbf{M}_{r_i}^\top\mathbf{e}_i, \mathbf{e}_j) = \frac{\mathbf{e}_i^\top \mathbf{M}_{r_i}\mathbf{e}_j}{\|\mathbf{M}_{r_i}^\top\mathbf{e}_i\|\|\mathbf{e}_j\|}.
\]
where $\mathbf{M}_{r_i}\in \mathbb{R}^{l\times l}$ and $\|\cdot\|$ is the L2-norm.
Instead of using $\mathbf{e}_i^\top \mathbf{M}_{r_i}\mathbf{e}_j$ as the score function, CODER uses $\cos$ to bound $S^{rel}_{ij}$.
When $\tau_{ij}=1$, $S_{ij}$ can be considered as a special case of $S_{ij}^{rel}$ with $\mathbf{M}_{r_i}=\mathbf{I}$.
The original DistMult samples corrupted triplets $(h',r,t')$ as negative training sample of $(h,r,t)$, while CODER utilizes $(h_i,r_i,t_{j-k})$ as negative term-relation-term pairs in a mini-batch (recall that $k+1\leq j\leq 2k$ and that the index for tail entities is 1 to $k$).

The term-relation-term pair label $\tau^{rel}_{ij}$ is defined by:
\[
    \tau^{rel}_{ij} = \left\{
\begin{aligned}
1&, & (h_i,r_i,t_{j-k})\in {\rm KG}, \\
0&, & (h_i,r_i,t_{j-k})\notin {\rm KG}.
\end{aligned}
\right.
\]
It is time-consuming to check if a relation triplet belongs to the KG.
In practice, we calculate:
\[
    \tau^{rel}_{ij} = \left\{
\begin{aligned}
1&, & t_i = t_{j-k}, \\
0&, & t_i \neq t_{j-k}.
\end{aligned}
\right.
\]
With $S^{rel}_{ij}$ and $\tau^{rel}_{ij}$ defined, the MS-loss can also be used for learning knowledge embeddings.
CODER mines hard term-relation-term pairs for anchor $(s_i, r_i)$ via:
\[
    \mathscr{N}_i^{rel}\coloneqq\{j\vert\tau_{ij}^{rel}=0,S_{ij}^{rel}>\min_{\tau^{rel}_{ik}=1}S_{ik}^{rel}-\epsilon^{rel}\},
\]
\[
    \mathscr{P}^{rel}_i\coloneqq\{j\vert\tau_{ij}^{rel}=1,S_{ij}^{rel}<\max_{\tau_{ik}^{rel}=0}S_{ik}^{rel}+\epsilon^{rel}\},
\]
where $\epsilon^{rel}$ is a margin.
The MS-loss of term-relation-term pairs is:
\[
    \mathcal{L}_{MS}^{rel} = \frac{1}{k}\sum_{i=1}^{k}(\frac{\log(1+\sum_{j\in\mathscr{P}^{rel}_i}\exp(-\alpha^{rel}(S^{rel}_{ij}-\lambda^{rel})))}{\alpha^{rel}}
\]
\[
     +\frac{\log(1+\sum_{j\in\mathscr{N}^{rel}_i}\exp(\beta^{rel}(S_{ij}^{rel}-\lambda^{rel})))}{\beta^{rel}}),
\]
where $\alpha^{rel}, \beta^{rel}, \lambda^{rel}$ are hyper-parameters.
To avoid degeneration of $\mathscr{P}_i, \mathscr{N}_i, \mathscr{P}_i^{rel}, \mathscr{P}_i^{rel}$, we guarantee for each triplet $(h_i,r_i,t_i)$, there exists at least $m$ different indexes $\{i_o\}_{o=1}^m$ such that:
\[
    (h_i,r_i,t_i) = (h_{i_o},r_{i_o},t_{i_o}),
\]
where $2\leq m \leq \sqrt{k}$.

The total loss is formulated as:
\[
    \mathcal{L}=\mathcal{L}_{MS} + \mu\mathcal{L}_{MS}^{rel}.
\]
The parameters of CODER are updated to minimize $\mathcal{L}$.
The overall model structure of CODER is illustrated in Figure \ref{workflow}.

\section{Experiments}
In this section, we first introduce the UMLS and the training of CODER.
Then we introduce the compared embeddings and the benchmarks:
medical term normalization tasks, a medical similarity measure, and relation classification on concepts.

\subsection{UMLS Metathesaurus}
We leverage the UMLS to build the training dataset.
The UMLS Metathesaurus contains medical concepts integrated from different lexicon resources.
Each concept has a Concept Unique Identifier (CUI) with multiple synonymous names (terms) potentially in multiple languages.
For example, concept \textbf{C0002502} has terms: Amilorid, Amiloride, producto con amilorida, etc.
Each concept has been assigned one or occasionally multiple semantic types.
For example, the semantic types of \textbf{C0002502} are Organic Chemical and Pharmacologic Substance.
The UMLS also provides relation information between medical concepts in the form of triplets, such as (\textbf{C5190625}, \textbf{C0002502}, (``CHD'', ``is a'')). ``CHD'' is the relation type and ``is a'' is the detailed relationship attribute in the UMLS.
Some triplets only contain the relation type or the relationship attribute.
We concatenate the names of the relation type and the attribute as relation labels of triplets.
CODER is trained using the UMLS 2020AA release, which contains 4.27 M concepts, 15.48 M terms, and 87.89 M relations with 127 semantic types, 14 relations types, and 923 relationship attributes.
Training samples are generated by relations and terms sampling according to \ref{method}.

\subsection{Training settings of CODER}\label{train}

We train ${\rm CODER_{ENG}}$ initialized from PubMedBERT for the English version with 100K training steps.
We also train cross-lingual version ${\rm CODER_{ALL}}$ initialized from mBERT with 1M training steps.
Terms are encoded by $\rm [CLS]$ representation.
A batch-size of $k=128$ relation triplets and gradient accumulation steps of 8 are used for training.
In each mini-batch, we set the count of repeat triplets $m=8$. 
The maximal sequence length is 32 since the terms are short.
We use AdamW \cite{loshchilov2017decoupled} as the optimizer with a linear warm-up in the first 10000 steps to a peak of 2e-5 learning rate that decayed to zero linearly.
The parameters $\alpha=\alpha^{rel},\beta=\beta^{rel},\lambda=\lambda^{rel},\epsilon=\epsilon^{rel}$ of MS-loss are set to 2, 50, 0.5, 0.1 respectively.
The weight $\mu$ is set to 1 to balance term-term loss and term-relation-term loss.

\subsection{Compared baseline embeddings}
We compare CODER with the following medical embeddings.
Details of compared embeddings are listed in {\bf Appendix A.1 Summary of compared medical embeddings}.

\textbf{Word Embedding} GoogleNews-vectors \cite{Mikolov2013} is a baseline of word embedding. Wiki-pubmed-PMC \cite{moen2013distributional}, BioNLP-2016-win-2 \cite{Chiu2016}, and BioNLP-2016-win-30 \cite{Chiu2016} are medical word embeddings with different training settings. To represent a term, we used the average of the embeddings of its words.

\textbf{Concept Embedding} Cui2vec \cite{beam2018clinical}, DeVine-etal-200 \cite{DeVine2014}, and claims-codes-hs-300 \cite{Choi20162} map different sets of medical concepts to vectors.
Cui2vec contains medical concepts from SNOMED-CT.
DeVine-etal-200 includes medical concepts that appeared in MedTrack and OHSUMED.
Claims-codes-hs-300 contains ICD-9 codes, LOINC laboratory codes, and NDC medication codes.

\textbf{Contextual Embedding} BERT-base-cased and mBERT \cite{devlin2018} are the baselines of contextual embedding.
BioBERT \cite{lee2020biobert} is the most well-known biomedical language model.
There exist two clinical language models that have the same name ClinicalBERT \cite{alsentzer2019publicly,huang2019clinicalbert}, and we distinguish them by the author’s name. 
We compare SciBERT \cite{beltagy-etal-2019-scibert}, BlueBERT \cite{peng2019transfer} and PubMedBERT \cite{gu2020domain} which are biomedical specific language models.
We also compare SapBERT \cite{sapbert}, which is a BERT-based medical term representation trained using contrastive learning on the terms of the UMLS.
We calculate two representations for contextual embedding $\mathbf{e}$ and $\mathbf{e}_{avg}$ for a term as we described in Section~\ref{rep}.

\subsection{Medical term normalization tasks}

We evaluate term normalization on three datasets in different languages.
Term normalization tasks are evaluated in a zero-shot setting, i.e., no training datasets are provided.

\subsubsection{Cadec, PsyTar}

Cadec \cite{Karimi2015} and PsyTar \cite{Zolnoori2019} are two English medical term normalization datasets.
Cadec contains 6754 terms and 1029 standard terms as normalization targets.
PsyTar contains 6556 terms, which are mapped to 618 standard terms.
We use the data splittings from \cite{Tutubalina2018, Miftakhutdinov} and
top-$k$ accuracy as the metric.
We report the average scores on different test sets.

Results on Cadec and PsyTar are shown in Table~\ref{cadec psytar}.
Appendix \ref{appendix_normalization} shows the results of contextual embeddings using $\rm [CLS]$ representation. 
For both datasets, CODER achieved superior performance over other baseline embeddings.
$\rm CODER_{ENG}$ based on KG contrastive learning outperformed SapBERT  that only uses terms by 5.72, 4.82, 1.47, 5.08
in terms of Cadec acc@1, acc@3, PsyTar acc@1, and acc@3, respectively.
$\rm CODER_{ALL}$ is trained by multiple languages and this may have caused $\rm CODER_{ALL}$ to perform slightly worse than $\rm CODER_{ENG}$.
We observe two interesting phenomena:
(1) Contextual embeddings pre-trained by masked language models fail to achieve embedding-based normalization.
(2) Medical word embeddings do not improve the word embedding baseline.

\begin{table}[h]
\caption{\label{cadec psytar} Acc@k for different embeddings in Cadec and PsyTar datasets. Contextual embeddings report results using the average pooling representation.
  }
\begin{tabular}{ccccc}
\toprule
 \multirow{2}{*}{Embedding} &
 \multicolumn{2}{c}{Cadec} &
 \multicolumn{2}{c}{PsyTar} \\
 \cmidrule(lr){2-3}
 \cmidrule(lr){4-5}
  & acc@1 & acc@3 & acc@1 & acc@3 \\
 \midrule
  GoogleNews-vectors&39.15&52.41&29.75&47.52 \\
  Wiki-pubmed-PMC&35.04&47.14&23.97&36.02 \\
  BioNLP-2016-win-2&35.47&48.30&25.29&37.70 \\
  BioNLP-2016-win-30&38.78&51.64&28.27&43.26 \\
 \midrule
 BERT-base-cased&19.64&25.65&15.16&19.84\\
 BioBERT&3.62&17.62&7.64&10.98 \\
 ClinicalBERT(Huang)&17.04&23.25&13.97&19.40 \\
 ClinicalBERT&13.77&18.40&11.12&15.47\\
 SciBERT&12.50&17.95&11.04&15.12 \\
 BlueBERT&14.64&20.62&13.24&18.07 \\
 PubMedBERT&11.50&15.88&9.76&13.78 \\
 SapBERT&54.05&71.42&52.45&66.04 \\
 $\rm CODER_{ENG}$&\textbf{59.77}&\textbf{76.24}&\textbf{53.92}&\textbf{71.12}\\
 $\rm CODER_{ALL}$&56.02&72.11&43.86&60.20\\
 \bottomrule
\end{tabular}
\end{table}

\subsubsection{The MANTRA GSC}
The MANTRA GSC \cite{Kors2015} contains French, German, Spanish, and Dutch corpus and includes medical terms mapped to a subset of the UMLS from three different corpora (MEDLINE, EMEA, and Patent).
To our knowledge, only part of these non-English corpora have been evaluated on before.
We compare $\rm CODER_{ALL}$ with two translation-based methods: biomedical translation model (BTM) \cite{roller2018cross} and a combination of Google Translator and Bing Translator (GB) \cite{roller2018cross}. 
We also compare with mBERT as a multi-lingual contextual embedding baseline.

$\rm CODER_{ALL}$ and mBERT search standard terms in SNOMED-CT, MeSH, MedDRA of UMLS2020AA including 801038 concepts and return one CUI per input term according to the cosine similarity.
We use F1-score as the metric since one term can map to multiple CUIs in the MANTRA GSC. 

\begin{table}
\caption{\label{mantra} F1-score of the MANTRA GSC for translation and embedding-based methods.}
\begin{tabular}{ccccccccccc}
 \toprule
 Method&SPA&FRE&DUT&GER  \\
 \midrule
 &\multicolumn{4}{c}{MEDLINE}\\
 BTM&69.1&67.4&61.4&66.3\\
 GB&68.7&\textbf{68.6}&\textbf{64.8}&67.9\\
 mBERT$\rm (\mathbf{e})$&13.6&16.7&16.3&17.3\\
 mBERT$\rm (\mathbf{e}_{avg})$&20.6&19.8&17.1&21.4\\
 $\rm CODER_{ALL}(\mathbf{e})$&70.1&58.6&58.6&\textbf{69.0}\\
 $\rm CODER_{ALL}(\mathbf{e}_{avg})$&\textbf{70.4}&58.9&59.5&68.4\\
 \midrule
 &\multicolumn{4}{c}{EMEA}\\
 mBERT$\rm (\mathbf{e})$&17.9&25.5&16.3&14.3&\\
 mBERT$\rm (\mathbf{e}_{avg})$&22.2&29.0&20.1&18.4&\\
 $\rm CODER_{ALL}(\mathbf{e})$&\textbf{68.1}&\textbf{62.9}&61.7&65.3\\
 $\rm CODER_{ALL}(\mathbf{e}_{avg})$&\textbf{68.1}&61.1&\textbf{62.9}&\textbf{65.6}\\
 \midrule
 &\multicolumn{4}{c}{Patent}\\
 mBERT$\rm (\mathbf{e})$&-&25.6&-&19.8\\
 mBERT$\rm (\mathbf{e}_{avg})$&-&32.4&-&24.3\\
 $\rm CODER_{ALL}(\mathbf{e})$&-&70.8&-&69.0\\
 $\rm CODER_{ALL}(\mathbf{e}_{avg})$&-&\textbf{71.1}&-&\textbf{70.2}\\
 \bottomrule\\
\end{tabular}
\end{table}

Table~\ref{mantra} summarizes the results of MANTRA.
$\rm CODER_{ALL}$ benefits from the UMLS multi-language synonyms, and has made obvious improvement over mBERT.
BTM is trained using additional dictionary and GB are trained on massive parallel corpus and achieve better performance than $\rm CODER_{ALL}$ in the French and Dutch MEDLINE.
$\rm CODER_{ALL}$ achieves state-of-the-art performance in the Spanish and Germany MEDLINE, EMEA, and Patent corpora.

\subsection{MCSM}
\citet{Choi20162} introduce Medical Conceptual Similarity Measure (MCSM) for evaluating medical embeddings.
MCSM is defined by:
\[
    {\rm MCSM}(V,T,k)=\frac{1}{|V(T)|}\sum_{v\in V(T)}\sum_{i=1}^k\frac{1_T(v(i))}{\log_2(i+1)},
\]
where $V$ is a set of medical concepts, $T$ is a semantic type, $|V(T)|$ is the number of medical concepts with semantic type $T$, $k$ is neighborhood size, $v(i)$ is the $i^{th}$ closest neighbor, and $1_T(\cdot)$ is the indicator function of semantic type $T$.
We re-implement the evaluation code of MCSM to support word and contextual embeddings.

We use SNOMED-CT as $V$ which contains 157726 concepts. 
Following the setting of \cite{Choi20162}, Pharmacologic Substance (PS), Disease or Syndrome (DS), Neoplastic Process (NP), Clinical Drug (CD), Finding (FD), and Injury or Poisoning (IP) are evaluated semantic types $T$ and $k$ is set to 40.

\begin{table}[h]
\caption{\label{mcsm} MCSM scores for different embeddings. Contextual embeddings report results with average pooling representation.}
\begin{tabular}{ccccccc}
 \toprule
    Embedding&PS&DS&NP&CD&FD&IP\\
    \midrule
    cui2vec&4.2&7.2&7.6&4.3&7.2&8.4\\
    DeVine-etal-200&7.3&6.9&6.3&1.2&4.6&2.5\\
    claims-codes-hs-300&9.4&7.7&7.5&-&3.4&8.6\\
    \midrule
    GoogleNews-vectors&9.1&9.3&9.6&10.0&8.5&10.3\\
    Wiki-pubmed-PMC&8.6&9.4&10.0&9.8&8.9&10.3 \\
    BioNLP-2016-win-2&9.1&9.4&10.0&10.2&8.9&\textbf{10.4} \\
    BioNLP-2016-win-30&8.3&9.3&10.0&9.6&8.8&10.3 \\
    \midrule
    BERT-base-cased&8.6&8.4&8.7&9.7&8.4&9.3\\
    BioBERT&8.7&8.8&9.3&9.5&7.8&9.3\\
    ClinicalBERT(Huang)&8.3&7.5&7.8&9.7&7.5&9.3\\
    ClinicalBERT&8.4&8.5&8.9&9.6&7.5&9.6\\
    SciBERT&7.6&6.9&6.2&8.5&6.3&7.5\\
    BlueBERT&8.1&7.4&8.5&9.3&7.4&9.2\\
    PubMedBERT&9.4&9.4&9.5&10.0&8.7&10.2\\
    SapBERT&8.2&8.3&8.5&7.9&8.6&9.1\\
    $\rm CODER_{ENG}$&\textbf{10.2}&9.4&9.4&\textbf{10.7}&9.6&10.1\\
    $\rm CODER_{ALL}$&\textbf{10.2}&\textbf{9.5}&\textbf{10.1}&10.6&\textbf{9.8}&10.3\\
  \bottomrule
\end{tabular}
\end{table}

MCSM results are reported in Table~\ref{mcsm}, where terms are represented by average pooling for contextual embeddings.
Appendix \ref{appendix_mcsm} lists MCSM results using $\rm [CLS]$ representation.
Concept embeddings suffer from OOV, which cannot cover all SNOMED-CT concepts and perform poorly on MCSM.
Though CODER is not designed for semantic classification or clustering, it still learns semantic information via relations from the UMLS and achieves state-of-the-art in five categories.
CODER is only bettered by word embeddings with a small margin in IP.

\subsection{Diseases Database relation classification}
We introduce the Diseases Database relation classification (DDBRC) task to classify the relation type using the head and tail concept embeddings.
131,741 medical relation triplets are collected from the Diseases Database \footnote{\url{http://www.diseasesdatabase.com/}}.
A triplet contains a head concept, a tail concept, and a relationship that can be classified into 14 categories.
18,419 of the head and tail concept pairs overlapped with the UMLS (though the concepts may have different recorded relation types).
We split the data 4:1 randomly as training and test sets.

Different embeddings are evaluated with the same classification model.
Denote the embedding of the head concept as $\mathbf{e_h}$ and that of the tail concept as $\mathbf{e_t}$.
The relation is predicted by:
\[
    \mathbf{\hat{y}} = softmax(W[\mathbf{e_h},\mathbf{e_t}] + \mathbf{b}).
\]
We use cross-entropy as the loss function.
Embeddings are tested as fixed (feature-based) or trainable (fine-tuned).
Further training details are listed in Appendix \ref{appendix_ddbrc}.
The classification accuracy on the test set is used as the metric.

Results of DDBRC are listed in Table~\ref{diseasedb}.
Contextual embeddings are powerful when fine-tuned.
Whether embeddings are trainable or not, CODER always achieves the highest accuracy.
Under the feature-based setting, $\rm CODER_{ENG}$ improves
7.1, 4.05, 4.25 over the best of concept, word, and contextual embeddings, respectively.
CODER also has the minimum performance difference between feature-based and fine-tuned among all contextual embeddings, and this indicates that the medical knowledge encoded into CODER makes them informative machine learning features.

\begin{table}[htbp]
\caption{\label{diseasedb} Testing accuracy of DDBRC for training 50 epochs.}
\begin{tabular}{cccc}
  \toprule
  Embedding&feature&fine-tune \\
  \midrule 
    cui2vec&88.95&88.92\\
    DeVine-etal-200&80.42&80.44\\
    claims-codes-hs-300&72.02&71.99\\
    \midrule 
    GoogleNews-vectors&90.96&90.85\\
    Wiki-pubmed-PMC&91.67&91.16\\
    BioNLP-2016-win-2&91.94&92.00\\
    BioNLP-2016-win-30&92.00&91.46\\
    \midrule 
    BERT-base-cased&78.22&98.74\\
    BioBERT&83.47&98.73\\
    ClinicalBERT(Huang)&75.82&98.77\\
    ClinicalBERT&76.12&98.76\\
    SciBERT&77.79&98.80\\
    BlueBERT&79.70&98.78\\
    PubMedBERT&75.93&98.81\\
    SapBERT&91.80&98.79\\
    $\rm CODER_{ENG}$&\textbf{96.05}&\textbf{98.85}\\
    $\rm CODER_{ALL}$&95.82&98.81\\
  \bottomrule 
\end{tabular}
\end{table}

\section{Discussions}

\subsection{Ablation study}
To explore how term-relation-term similarity impacts the normalization performance, we first train CODER without term-relation-term similarity.
We further test another TransE-like term-relation-term similarity function $S^{rel,T}_{ij} = -\|\mathbf{e}_{i}+\mathbf{r}_i-\mathbf{e}_{j}\|$.
Details of pre-training parameters are listed in Appendix \ref{appendix_ablation}.

\begin{table}[h]
\caption{\label{ablation} Ablation results for different training settings of CODER.}
\begin{tabular}{lcccc}
  \toprule
  \multirow{2}{*}{Setting}&\multicolumn{2}{c}{Cadec}&\multicolumn{2}{c}{PsyTar}\\
  \cmidrule(lr){2-3}
  \cmidrule(lr){4-5}
  &acc@1&acc@3&acc@1&acc@3\\
  \midrule
  $\rm CODER_{ENG}$&\textbf{59.77}&\textbf{76.24}&\textbf{53.92}&\textbf{71.12}\\
  $\ -S^{rel}_{ij}$&54.18&71.40&50.58&67.22\\
  $\ -S^{rel}_{ij}+S^{rel,T}_{ij}$&57.35&73.61&51.35&69.23\\
  \bottomrule
\end{tabular}
\end{table}

Table~\ref{ablation} shows ablation results on dataset Cadec and PsyTar.
Term normalization performance gains from relation information of the UMLS.
$S^{rel,T}_{ij}$ utilizes relation information of the UMLS, but it has an unbounded range and is inconsistent with the form of $S_{ij} = \cos(\mathbf{e}_i, \mathbf{e}_j)$.
Therefore, using $S^{rel}_{ij}$ is a better choice to encode relational knowledge with better normalization performance.


\subsection{Embedding visualization analysis}
\begin{figure}[h]
\centering
\includegraphics[scale=0.33]{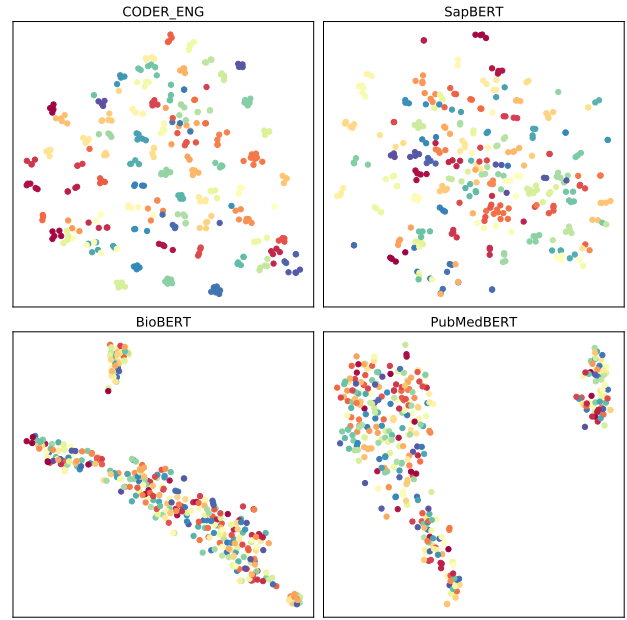}
\caption{\label{plot2} The t-SNE Visualization of different contextual embeddings for term normalization.}
\Description{The t-SNE Visualization of different contextual embeddings for term normalization.}
\end{figure}

\begin{figure}[h]
\centering
\includegraphics[scale=0.33]{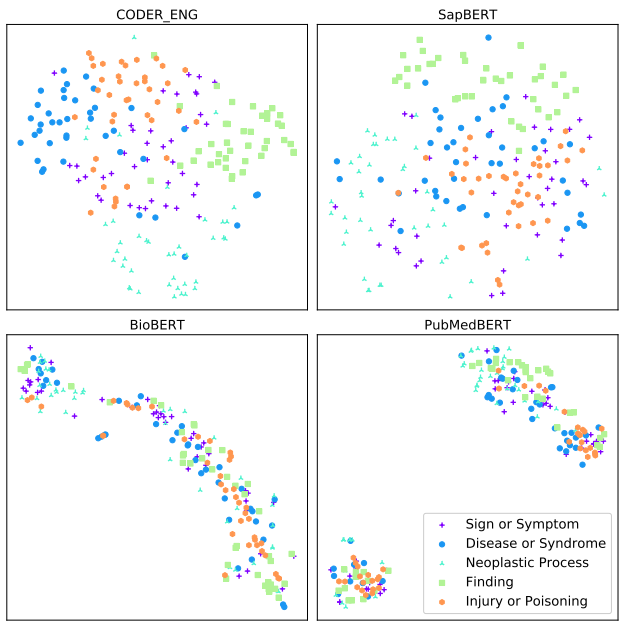}
\caption{\label{plot3} The t-SNE Visualization of different embeddings for concepts from different semantic types.}
\Description{The t-SNE Visualization of different embeddings for concepts from different semantic types.}
\end{figure}

We use t-SNE \cite{maaten2008visualizing} to visualize different medical embeddings.
Figure~\ref{plot2} and Figure~\ref{plot3} plot medical terms from the International Classification of Diseases ICD-10-CM.
Terms attached to the same concept are in the same color for Figure~\ref{plot2} and concepts are color-coded by semantic types in Figure~\ref{plot3}.
They show that $\rm CODER_{ENG}$ and SapBERT cluster terms automatically and group concepts according to semantic types, while BioBERT and PubMedBERT are incapable of clustering.
Notice that $\rm CODER_{ENG}$ can distinguish concepts more clearly.
SapBERT mixes up Sign or Symptom, Disease or Syndrome, and Injury or Poisoning a little, while $\rm CODER_{ENG}$ achieves better category clustering.

\subsection{Case study of CODER}

\subsubsection{Cross-lingual}
To illustrate the ability of $\rm CODER_{ALL}$ on cross-lingual term normalization, we search the closest English term to the terms in different languages for concept C0004604 ``\textit{dorsalgia}'' as a case study, as there is no medical normalization dataset for multiple languages.
We search for the most similar English terms by the cosine distance among all ICD-9-CM concepts, and the results are showed in Table~\ref{case}.
Most letter-based languages normalize correctly to the term ``\textit{dorsalgia}'' and character-based language (Chinese, Japanese, and Korean) did not perform as well.
Letter-based languages benefited from sharing similar sub-words like ``\textit{dolor}''.
The UMLS does not include Chinese terms, but CODER could normalize ``\begin{CJK*}{UTF8}{gbsn}背痛\end{CJK*}''\ because Chinese shares characters with Japanese.
Unfortunately, $\rm CODER_{ALL}$ could not handle the unique alphabets of Japanese and Korean well.

\begin{table}[h]
\caption{\label{case} The closest term of C0004604(dorsalgia) in different languages.}
\begin{tabular}{ccc}
  \toprule
  Language&Term&Target \\
  \midrule 
  Spanish&dolor de espalda&dorsalgia\\
  Hungarian&H\'atf\'aj\'as&dorsalgia\\
  Japanese&\begin{CJK*}{UTF8}{gbsn}背部痛\end{CJK*}&dorsalgia\\
  Japanese&\begin{CJK*}{UTF8}{gbsn}はいぶつう\end{CJK*}&puberty\\
  Czech&bolesti zad&dorsalgia\\
  Germany&R\"uckenschmerzen&dorsalgia\\
  Italian&Dolore dorsale&dorsalgia\\
  Korean&\begin{CJK*}{UTF8}{mj}배통\end{CJK*}&benign neoplasm of ureter\\
  Finnish&Selk\"akipu&dorsalgia\\
  French&Dorsalgie&dorsalgia\\
  Dutch&Pijn, rug-&dorsalgia\\
  Norwegian&Ryggsmerter&dorsalgia\\
  Croatian&BOL U LE\DJ IMA&dorsalgia\\
  Swedish&Ryggsm\"arta&dorsalgia\\
  Chinese&\begin{CJK*}{UTF8}{gbsn}背痛\end{CJK*}&dorsalgia\\
  \bottomrule
\end{tabular}

\end{table}

\subsubsection{Knowledge-injected term normalization}
To show knowledge can enhance the ability of CODER, we compare the most similar terms extracted by $\rm CODER_{ENG}$ and $\rm CODER_{ENG}$$-S^{rel}_{ij}$.


\begin{table}[h]
\caption{\label{knowledge} Top 5 similar terms searched from ICD-9-CM.}
\begin{tabular}{ccc}
  \toprule
  Top-k&pufferfish poisoning&angina\\
  \midrule
  &\multicolumn{2}{c}{$\rm CODER_{ENG}$}\\
  \midrule
  1&\textbf{toxic eff fish/shellfish}&\textbf{angina pectoris}\\
  2&poisoning by saluretics&\textbf{chest pain}\\
  3&toxic effect of fusel oil&prinzmetal angina\\
  4&poisoning by paraldehyde&vincent's angina\\
  5&poisoning by succinimides&angina decubitus\\
  \midrule
  &\multicolumn{2}{c}{$\rm CODER_{ENG}$$-S^{ij}_{rel}$}\\
  \midrule
  1&poisoning by saluretics&\textbf{angina pectoris}\\
  2&poison antivaricose drug&vincent's angina\\
  3&poisoning by paraldehyde&prinzmetal angina\\
  4&anphylct reaction fish&angina decubitus\\
  5&louping ill&pois coronary vasodilat\\
 \bottomrule
\end{tabular}
\end{table}

We list top similar diagnosis terms for ``\textit{pufferfish poisoning}'' and ``\textit{angina}'' from ICD-9-CM in Table ~\ref{knowledge}.
With additional knowledge from the UMLS, $\rm CODER_{ENG}$ can normalize ``\textit{pufferfish poisoning}'' to ``\textit{toxic eff fish/shellfish}''.
Both $\rm CODER_{ENG}$ and $\rm CODER_{ENG}$$-S^{ij}_{rel}$ can normalize ``\textit{angina}'' to ``\textit{angina pectoris}''.
$\rm CODER_{ENG}$ further finds the ``\textit{chest pain}'' is similar to ``\textit{angina}'', which is informative since ``\textit{chest pain}'' may be caused by ``\textit{angina}''.

\section{Conclusions}

In this paper, we present a medical term embedding model CODER.
The KG contrastive learning framework learns both term-term similarity and term-relation-term similarity.
Evaluations show that CODER substantially outperforms existing medical embeddings in zero-shot term normalization and that it also exhibits good performance in semantic similarity and relation classification.
$\rm CODER_{ALL}$ is the first cross-lingual medical term embedding model, which can help low-resource languages in medical NLP tasks.

\bibliographystyle{ACM-Reference-Format}
\bibliography{main}

\newpage
\appendix

\section{Appendices}

\subsection{A summary of compared medical embeddings}

We list a summary of medical embeddings in Table~\ref{summary}.
Several versions of BioBERT, ClinicalBERT, SciBERT, BlueBERT, and PubMedBERT exist with different training corpora.

We choose BioBERT-base v1.1 \footnote{\url{https://drive.google.com/file/d/1R84voFKHfWV9xjzeLzWBbmY1uOMYpnyD}},
Bio\_ClinicalBERT \footnote{\url{https://huggingface.co/emilyalsentzer/Bio_ClinicalBERT}},
scibert-scivocab-uncased \footnote{\raggedright\url{https://s3-us-west-2.amazonaws.com/ai2-s2-research/scibert/huggingface_pytorch/scibert_scivocab_uncased.tar}},
BlueBERT-base-uncased-PumMed+MIMIC III \footnote{\raggedright\url{https://huggingface.co/bionlp/bluebert_pubmed_mimic_uncased_L-12_H-768_A-12}},
BiomedNLP-PubMedBERT-base-uncased-abstract \footnote{\raggedright\url{https://huggingface.co/microsoft/BiomedNLP-PubMedBERT-base-uncased-abstract}}
to compare.

\begin{table}[h]
\caption{\label{summary} Summary of compared medical embeddings.
  }
\begin{tabular}{cccc}
  \toprule
  Embedding&\# Words&Dim.&Category\\
  \midrule
    cui2vec \footnotemark[7]&108477&500&concept\\
    DeVine-etal-200 \footnotemark[8]&52102&200&concept\\
    claims-codes-hs-300 \footnotemark[9]&14852&300&concept\\
    \midrule
    GoogleNews-vectors \footnotemark[10]&3.0M&300&word\\
    Wiki-pubmed-PMC \footnotemark[11]&5.5M&200&word\\
    BioNLP-2016-win-2 \footnotemark[12]&2.2M&200&word\\
    BioNLP-2016-win-30 \footnotemark[13]&2.2M&200&word\\
    \midrule
    BERT-base-cased \footnotemark[14]&-&768&contextual\\
    BioBERT&-&768&contextual\\
    ClinicalBERT(Huang) \footnotemark[15]&-&768&contextual\\
    ClinicalBERT&-&768&contextual\\
    SciBERT&-&768&contextual\\ 
    BlueBERT&-&768&contextual\\
    PubMedBERT&-&768&contextual\\
    SapBERT \footnotemark[16]&-&768&contextual\\
  \bottomrule 
 \end{tabular}
\end{table}

\footnotetext[7]{\url{https://figshare.com/s/00d69861786cd0156d81}}
\footnotetext[8]{\url{https://github.com/clinicalml/embeddings/blob/master/DeVine_etal_200.txt.gz}}
\footnotetext[9]{\url{https://github.com/clinicalml/embeddings/blob/master/claims_codes_hs_300.txt.gz}}
\footnotetext[10]{\url{https://drive.google.com/file/d/0B7XkCwpI5KDYNlNUTTlSS21pQmM}}
\footnotetext[11]{\raggedright\url{http://evexdb.org/pmresources/vec-space-models/wikipedia-pubmed-and-PMC-w2v.bin}}
\footnotetext[12]{\url{https://drive.google.com/file/d/0BzMCqpcgEJgiUWs0ZnU0NlFTam8}}
\footnotetext[13]{\url{https://drive.google.com/file/d/0BzMCqpcgEJgiUWs0ZnU0NlFTam8}}
\footnotetext[14]{\raggedright\url{https://storage.googleapis.com/bert_models/2018_10_18/cased_L-12_H-768_A-12.zip}}
\footnotetext[15]{\url{https://drive.google.com/file/d/1t8L9w-r88Q5-sfC993x2Tjt1pu--A900}}
\footnotetext[16]{\url{https://huggingface.co/cambridgeltl/SapBERT-from-PubMedBERT-fulltext}}

\subsection{English terms normalization results for using [CLS] representation}\label{appendix_normalization}
Results of Cadec and PsyTar for contextual embeddings using $\rm [CLS]$ representation are shown in Table~\ref{norm_e}.
Compared to Table~\ref{cadec psytar}, we find using average pooling representation is beneficial for term normalization.

\begin{table}[h]
\caption{\label{norm_e} Acc@k for contextual embeddings in Cadec and PsyTar datasets. Contextual embeddings report results with $\rm [CLS]$ representation.
  }
\begin{tabular}{ccccc}
\toprule
 \multirow{2}{*}{Embedding} &
 \multicolumn{2}{c}{Cadec} &
 \multicolumn{2}{c}{PsyTar} \\
 \cmidrule(lr){2-3}
 \cmidrule(lr){4-5}
  & acc@1 & acc@3 & acc@1 & acc@3 \\
 \midrule
 BERT-base-cased&8.03&11.03&7.07&9.66\\
 BioBERT&8.33&11.58&4.36&6.38\\
 ClinicalBERT(Huang)&9.91&14.78&9.02&12.60\\
 ClinicalBERT&7.18&8.84&5.92&7.83\\
 SciBERT&10.61&14.66&8.79&12.55\\
 BlueBERT&11.42&15.13&9.08&12.10\\
 PubMedBERT&11.32&14.79&8.93&13.32\\
 SapBERT&49.26&62.61&44.22&58.62\\
 $\rm CODER_{ENG}$&\textbf{58.53}&\textbf{75.96}&\textbf{55.27}&\textbf{73.31}\\
 $\rm CODER_{ALL}$&52.93&66.53&38.36&52.48\\
 \bottomrule
\end{tabular}
\end{table}

\subsection{MCSM results for using [CLS] representation} \label{appendix_mcsm}
We list MCSM scores for contextual embeddings using $\rm [CLS]$ representation in Table~\ref{mcsm_e}.
Compared to Table~\ref{mcsm}, average pooling representation achieves better performance on semantic similar measurement except for CODER.

\begin{table}[h]
\caption{\label{mcsm_e} MCSM results for contextual embeddings using $\rm [CLS]$ representation.
  }
\begin{tabular}{ccccccc}
 \toprule
    Embedding&PS&DS&NP&CD&FD&IP\\
    \midrule
    BERT-base-cased&8.1&7.2&6.2&9.0&6.3&8.2\\
    BioBERT&8.6&8.3&8.5&9.2&6.9&9.0\\
    ClinicalBERT(Huang)&7.5&6.4&5.7&8.6&5.5&8.5\\
    ClinicalBERT&7.0&7.1&7.2&8.9&5.9&8.0\\
    SciBERT&7.6&6.9&6.2&8.5&6.3&7.5\\
    BlueBERT&7.0&6.3&6.4&8.4&5.9&7.7\\
    PubMedBERT&9.3&9.4&9.1&10.0&8.4&10.0\\
    SapBERT&9.4&8.8&9.3&9.7&9.1&9.8\\
    $\rm CODER_{ENG}$&10.3&\textbf{9.5}&9.6&\textbf{10.7}&9.5&\textbf{10.1}\\
    $\rm CODER_{ALL}$&\textbf{10.4}&\textbf{9.5}&\textbf{9.8}&\textbf{10.7}&\textbf{9.7}&\textbf{10.1}\\ 
  \bottomrule
\end{tabular}
\end{table}

\subsection{Training parameters of DDBRC}\label{appendix_ddbrc}
For feature-based, the learning rate is set at 1e-3 for all kinds of embeddings.
For fine-tuning, the learning rate is set at 1e-3 for word embeddings and concept embeddings and 2e-5 for contextual embeddings.
Batch-size is 512 for feature-based contextual embeddings, word embedding, and concept embeddings.
For fine-tuned contextual embeddings, batch-size is 96.
Contextual embeddings use $\mathbf{e}$ to represent terms.
All models are trained for 50 epochs.



\subsection{Ablation parameters}\label{appendix_ablation}

In the ablation study, all models are trained with 100K training steps and initialized from PubMedBERT.

We adopt different parameters of MS-loss since the TransE-like term-relation-term similarity function $S^{rel,T}_{ij}$ is unbounded.
The parameter $\alpha,\beta,\lambda,\epsilon$ of MS-loss are still set to 2, 50, 0.5, 0.1 respectively.
We notice that the L2-norm of term-relation-term similarity is close to $28$ when the model is initialized, and we let $\lambda^{rel}=28$ to avoid exponential function explode.
We also set $\alpha^{rel},\beta^{rel},\epsilon^{rel}$ to 0.2, 5, 0.1 for numerical stability.
The weight $\mu$ is set to 0.1 to balance term-term loss and term-relation-term loss.

All other parameters are the same with Section \ref{train}.

\end{document}